\documentclass[10pt,journal]{IEEEtran}

\ifCLASSOPTIONcompsoc
\else
\fi
\ifCLASSINFOpdf
\else
\fi
\usepackage{times}
\usepackage{epsfig}
\usepackage{graphicx}
\usepackage{amsmath}
\usepackage{color}
\usepackage{amssymb}
\usepackage{xcolor}
\usepackage[utf8]{inputenc} 
\usepackage[T1]{fontenc}    
\usepackage{url}            
\usepackage{booktabs}       
\usepackage{amsfonts}       
\usepackage{nicefrac}       
\usepackage{microtype}      
\usepackage{amsmath}
\usepackage{graphicx}

\makeatletter 
\newcommand{\Rmnum}[1]{\expandafter\@slowromancap\romannumeral #1@}
\makeatother

\begin{document}
\title{Quadruplet Network with One-Shot Learning for Fast Visual Object Tracking}
\author{Xingping Dong, Jianbing Shen,~\IEEEmembership{Senior Member,~IEEE}, Dongming Wu, Kan Guo, \\
Xiaogang Jin,~\IEEEmembership{Member,~IEEE}, and Fatih~Porikli,~\IEEEmembership{Fellow,~IEEE}
\IEEEcompsocitemizethanks{
\IEEEcompsocthanksitem This work was supported in part by
the Beijing Natural Science Foundation under Grant 4182056,
the Key Research and Development Program of Zhejiang Province under Grant 2018C03055,
and the National Natural Science Foundation of China under Grant 61732015,
the Australian Research Council's Discovery Projects funding scheme under grant DP150104645,
and the Fok Ying-Tong Education Foundation for Young Teachers.
Specialized Fund for Joint Building Program of Beijing Municipal Education Commission.
(Corresponding author: \textit{Jianbing Shen})
\IEEEcompsocthanksitem X. Dong, J. Shen and D. Wu are with Beijing
Laboratory of Intelligent Information Technology, School of Computer
Science, Beijing Institute of Technology, Beijing 100081, P. R. China. \protect 
(email: shenjianbing@bit.edu.cn)
\IEEEcompsocthanksitem K. Guo is with Alibaba Group, Hangzhou 311121, P. R. China.
(email: guokan.gk@alibaba-inc.com)
\IEEEcompsocthanksitem X. Jin is with State Key Lab of CAD$\&$CG,
Zhejiang University, Hangzhou, 310058, P. R. China.
(email: jin@cad.zju.edu.cn)
\IEEEcompsocthanksitem F. Porikli is with the Research School of Engineering, the Australian National University.
(email: fatih.porikli@anu.edu.au)
}
}

%


\IEEEcompsoctitleabstractindextext{%
\begin{abstract}
In the same vein of discriminative one-shot learning, Siamese networks allow recognizing an object from a single exemplar with the same class label. However, they do not take advantage of the underlying structure of the data and the relationship among the multitude of samples as they only rely on pairs of instances for training. In this paper, we propose a new quadruplet deep network to examine the potential connections among the training instances, aiming to achieve a more powerful representation. We design a shared network with four branches that receive multi-tuple of instances as inputs and are connected by a novel loss function consisting of pair-loss and triplet-loss. According to the similarity metric, we select the most similar and the most dissimilar instances as the positive and negative inputs of triplet loss from each multi-tuple. We show that this scheme improves the training performance. Furthermore, we introduce a new weight layer to automatically select suitable combination weights, which will avoid the conflict between triplet and pair loss leading to worse performance. We evaluate our quadruplet framework by model-free tracking-by-detection of objects from a single initial exemplar in several Visual Object Tracking benchmarks. Our extensive experimental analysis demonstrates that our tracker achieves superior performance with a real-time processing speed of 78 frames-per-second (fps).
\end{abstract}

}

\maketitle
\IEEEdisplaynotcompsoctitleabstractindextext

\IEEEpeerreviewmaketitle

\section{Introduction}\label{sec:intro}
In recent years, deep learning models have attracted significant attention thanks to their powerful regression capacity in a spectrum of applications from speech recognition to natural language processing and computer vision. It is recognized that training of deep neural networks requires a large corpus of labeled data to attain generality of the learned models and robustness of the feature maps. Such networks seem less useful for one-shot learning tasks where the objective is to learn a model, often in an online fashion, from a single exemplar (or a few). One exemption is the embedding with Siamese deep networks \cite{bromley1993signature, bertinetto2016learning, bertinetto2016fully}, since it is not necessary to retrain the deep model for a newly given object or class. Siamese architectures can identify other instances of the target class from its original exemplar using a fixed model.

\begin{figure}
  \centering
	\includegraphics[width = .46\textwidth]{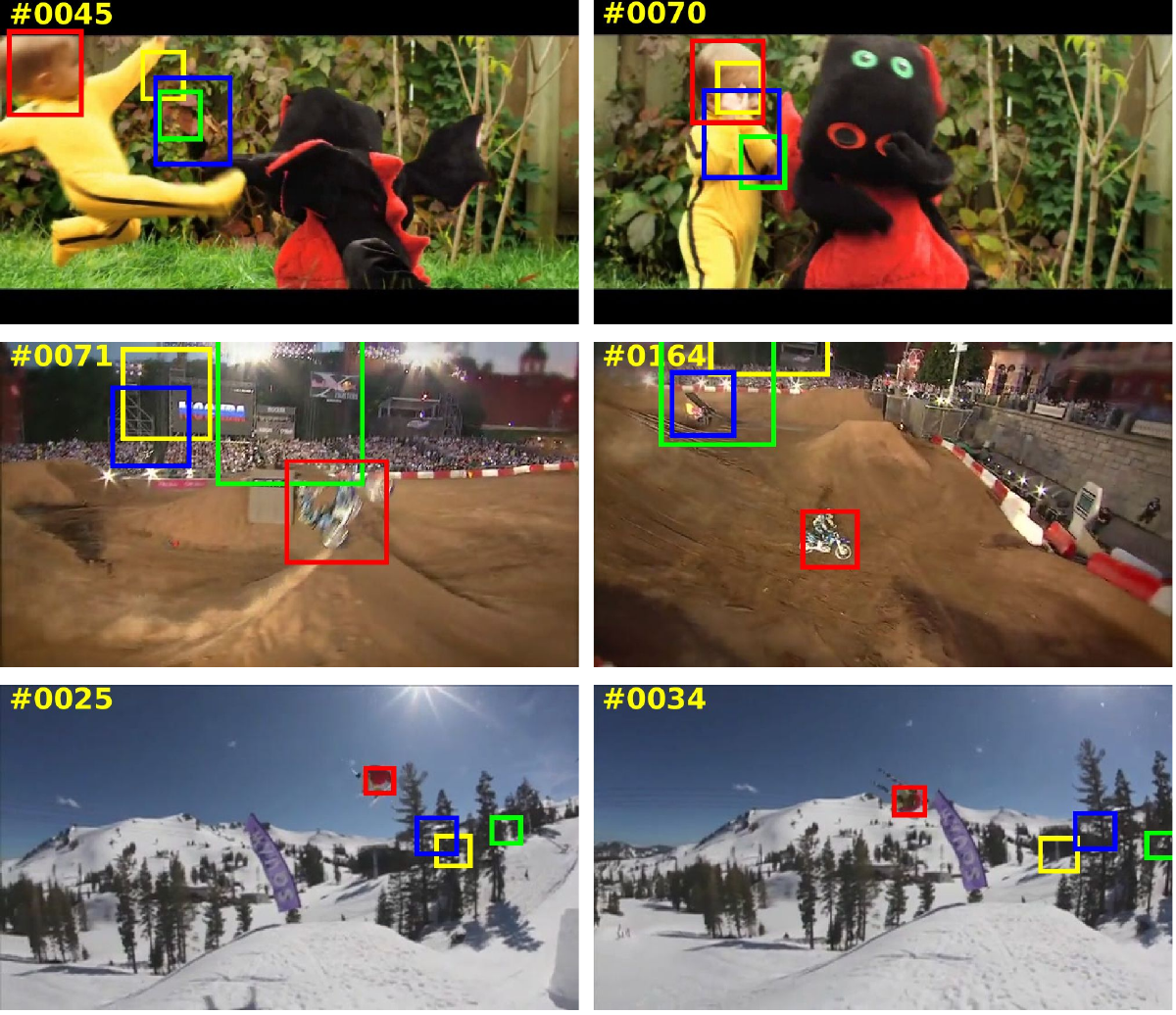}\\
    Staple \textcolor{yellow}{ \rule[0.3ex]{6mm}{1mm}}  CFnet-conv2 \textcolor{blue}{ \rule[0.3ex]{6mm}{1mm}} SiamFc-3s \textcolor{green}{ \rule[0.3ex]{6mm}{1mm}}    Ours \textcolor{red}{ \rule[0.3ex]{6mm}{1mm}}
  \caption{Sample results of our quadruplet network for visual tracking.
By only changing the training loss on base Siamese network \cite{bertinetto2016fully} to get more powerful feature, we achieve a significant improvement on the object tracking task.
For instance, our method improves the distance precision \cite{wu2013online} from $30.1\%$ to a higher score of $86.7\%$ on \textit{DragonBaby} (top row). It also outperforms the referenced real-time trackers Staple \cite{bertinetto2016staple} and CFnet-conv2 \cite{valmadre2017end}.}
  \label{example}
\end{figure}

Conventional Siamese networks use tuples of two labeled instances for training \cite{bromley1993signature, fan2014learning, lin2015bilinear, bertinetto2016learning, bertinetto2016fully}. They are sensitive to calibration and the adopted notion of similarity vs. dissimilarity depending on the given context \cite{hoffer2015deep}. The requirement of calibration can be removed by applying triplets for training with a contrastive loss, which favors a small distance between pairs of exemplars labeled as similar, and a large distance for pairs labeled dissimilar \cite{hoffer2015deep}. At the same time, triplets enable utilization of the underlying connections among more than two instances. Our intuition is that more instances (larger tuples) lead to better performance in the learning process. Therefore, we design a new network structure by adding as many instances into a tuple as possible (including a triplet and multiple pairs) and connect them with a novel loss combining a pair-loss and a triplet based contractive-loss. Existing Siamese \cite{bertinetto2016fully} or triplet networks \cite{hoffer2015deep} do not use the full potential of the training instances since they take randomly sampled pairs or triplets to construct the training batches. In contrast, our framework aims to select the triplets that would lead to a powerful representation for a stronger deep network.

In this paper, we introduce a novel quadruplet network for one-shot learning. Our quadruplet network is a discriminative model to one-shot learning. Given a single exemplar of a new object class, its learned model can recognize objects of the same class. To capture the underlying relationships of data samples, we design four branch networks with shared weights for different inputs such as instances, exemplar, positive and negative branches. We randomly sample a set of positive and negative instances as inputs of instances branch. A pair of loss function is designed to connect the exemplar and instances branch to utilize the underlying relationships of pairs. To achieve the triplet of representation, we select the powerful positive instance (which is most similar to the exemplar) and the negative instance (which is most dissimilar one) as the inputs of positive and negative branches, respectively. Then, we use a contractive loss function to measure the similarity of this triplet. Finally, the weighted average of the pair loss and the contractive loss is assigned as the final loss.

The triplet loss is the key to utilize the underlying connections among instances to achieve improved performance. To combine it and pair loss, a simple solution is to apply a weighted average with prior weights between these two losses. However, directly applying prior weights maybe not improve even reduce performance. For example, we test our approach with prior weights for visual object tracking on OTB-2013 benchmark \cite{wu2013online} while the distance precision is reduced from 0.809 to 0.800. Thus, we propose a weight layer to automatically choose suitable combination weights during training to solve this problem.
We evaluate the quadruplet network with one-shot learning for visual object tracking. SiamFc-3s \cite{bertinetto2016fully} is our baseline tracker. We apply our quadruplet network instead of its Siamese network to train the shared net and adapt the same mechanism for online tracking.
As shown in Fig. \ref{example}, our training method achieves better tracking accuracy, which demonstrates the more powerful representation of our framework.
In several popular tracking benchmarks, our experimental results show that the tracker runs at high real-time speed (78 frames-per-second on OTB-2013) and achieves excellent results compared with recent state-of-the-art real-time trackers.

The main contributions of this work are summarized as:
\begin{itemize}
\item We propose a novel quadruplet network for one-shot learning that utilizes the inherent connections among multiple instances and apply it to visual tracking. To the best of our knowledge, we are the first to introduce quadruplet network into single object tracking.
\item A weight layer is proposed for selecting suitable combination weights between triplet and pair loss. It may take a lot of time to manually choose appropriate combination weights while our weight layer is able to automatically adjust these weights during each training iteration to achieve powerful features.
\item By applying the proposed quadruplet network on training to get more representable features, our detection-by-tracking method can achieve state-of-the-art results even without online updating during tracking. Furthermore, our tracker runs beyond real-time with high speed of 78 fps.
\end{itemize}
%

\section{Related Work}\label{sec:related}
Our work is related to a wide array of literature, in particular, one-shot learning with generative models, learning an embedding space for one-shot learning, and visual object tracking.

Many previous studies focus on the context of the generative models for one-shot learning, which is different from our formulation of the problem as a discriminative task. One early approach \cite{fei2006one} uses probabilistic generative models to present object categories and applies a variational Bayesian framework for learning useful information from a handful of training samples. In \cite{rezende2016one},  a recurrent spatial attention generative model is proposed to generate images using a variational Bayesian inference. Having seen samples once, this method generates a set of diverse samples.

The most common discriminative approach to one-shot learning is embedding learning, where the goal is to learn an embedding space and then perform classification by applying a simple rule in the embedding space such as finding the nearest-neighbor of an exemplar of a novel category. A typical model is Siamese network \cite{bromley1993signature}. Recently, several techniques \cite{bertinetto2016learning, bertinetto2016fully, hoffer2015deep} are presented to improve the embedding performance. In \cite{bertinetto2016learning} a learning-to-learn approach is proposed to determine the shared weights of Siamese network. It applies a number of factorizations of the parameters to make the construction feasible. In \cite{bertinetto2016fully}, a fully convolutional Siamese network is designed for visual object tracking. As an alternative to using pairs in Siamese network, Hoffer \textit{et al.} \cite{hoffer2015deep} employs triplets for training by distance comparisons. In comparison, our method combines the pairs and triplets for training to take the advantage of the underlying useful connections (e.g. the manifold structure) among samples.

\begin{figure*}[ht]
  \centering
   \includegraphics[width = .88\textwidth]{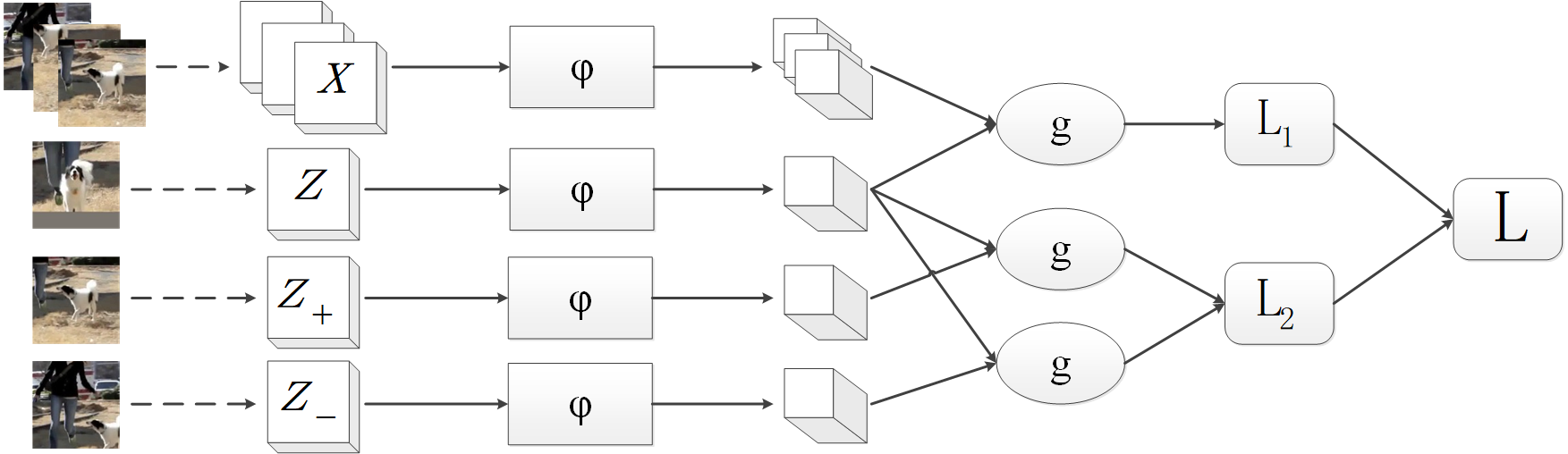}
  \caption{The structure of a quadruplet network. $x,z,z_+,z_-$ are different inputs of four shared branch networks respectively corresponding to instances, exemplar, positive and negative branches. $\phi$ represents a convolution network with shared weights. $g$ is a function of similarity metric. $L_1$, $L_2$, and $L$ are loss functions. Note that the $L_1$ loss is derived from exemplar and instances, while $L_2$ is from exemplar, positive instance and negative instance. $L$ loss is the weighted combination of these two losses. We further give more images in tracking task for illustration at the left of this figure.
  }
  \label{net}
\end{figure*}

The task of visual object tracking can be viewed as an application of one-shot learning \cite{bertinetto2016learning}. We only provide a brief introduction of recent object tracking methods such as correlation filter based trackers and deep neural networks based solutions. After Kernelized Correlation Filters \cite{Henriques2015High} had reported state-of-the-art performance at hundreds of frame-per-second speed, more investigation focused on correlation filters \cite{danelljan2014accurate,ma2015long,liu2015real,hong2015multi,bertinetto2016staple,DanelljanECCV2016,Cheng2018,Ma2016,Ma2015Hierarchical,Dong2017}. The tracking performance improved consistently, and DSST \cite{danelljan2014accurate} and C-COT \cite{DanelljanECCV2016} obtained the top ranks at the VOT 2014 \cite{Kristan2014a} and the VOT 2016 challenges \cite{Kristan2016a}, respectively. However, the speed became lower and lower, i.e. the speed of C-COT is only 1 fps. The computational load remains as the same problem in deep neural networks for tracking. The online learning network \cite{nam2015learning} provided superior tracking performance, albeit its prohibitive load (near 1 fps on GPU) limits its practicality. To accelerate deep network for tracking, some studies proposed off-line training of models, such as DeepTrack \cite{HanxiLi2014}, GOTURN \cite{held2016learning} and SiamFc-3s \cite{bertinetto2016fully}, and directly applied them for online tracking to avoid the cost of online retraining. All of these achieve a speed of more than real-time (30 fps) with a comparable performance. Our work is similar to SiamFc-3s, while we focus on improving off-line training phase to obtain a more robust feature representation. Thus, our method runs at a comparable speed with SiamFc-3s yet achieves better performance.

Recently, a lot of real-time (more than 30 fps) methods \cite{bolme2010visual,Henriques2015High,danelljan2014accurate,danelljan2014adaptive,bertinetto2016staple,choi2016visual,bertinetto2016fully,valmadre2017end,wang2017large,kiani2017learning,huang2017learning,dong2018hyperparameter,dong2018triplet} and nearly real-time (more than 20 fps) methods \cite{danelljan2014accurate,ma2015long,fan2017parallel,guo2017learning} are proposed for visual object tracking.
Our tracker can operate at high speed with 78 fps. Thus, we only introduce some trackers with high frame-rates (more than 50 fps) for reference. What is worth mentioning that all the following speeds are from the original paper, and are not tested on the same platform.

In the aspect of deep learning, Beritinetto \textit{et al.} \cite{bertinetto2016fully} apply 5 fully convolutional layers to train an end-to-end Siamese network (SiamFc-3s) for tracking, and directly use it for online tracking without updating frame by frame.
So it achieves high frame-rates beyond real-time, nearly 86 fps (78 fps in our machine) with GPU.
In CFnet \cite{valmadre2017end}, Valmadre \textit{et al.} embed a correlation filter layer into a Siamese network to enable learning deep features which are tightly coupled to the Correlation Filter (CF). It shows that 2 convolutional layers adding CF layer in Siamese network will achieve comparable performance and speed (76 fps) compared with SiamFc including 5 convolutional layers. Zhang \textit{et al.} \cite{zhang2017deep} learns an early decision  policy for different frames to speed up the SiamFc tracker and achieve a great improvement in terms of speed (159 fps), by using reinforcement learning.

In correlation filter trackers, the early MOSSE \cite{bolme2010visual} and improved Kernelized Correlation Filter (KCF) \cite{Henriques2015High} trackers are able to run at the frame-rate with 669 fps and 292 fps, respectively. The following CN \cite{danelljan2014adaptive} tracker combined with adaptive color name feature can achieve speed at 105 fps. Another work Staple \cite{bertinetto2016staple} makes attempt to use multiple features containing color and HOG to get robust performance and high speed (80 fps). Recently, Wang \textit{et al.} \cite{wang2017large} combine the correlation filter algorithm and accelerate the Structured output support vector machine (SVM) based tracker using the correlation filter algorithm, and it can also run at 80 fps.

\section{Siamese Networks for Tracking}\label{sec:siam}
Here, we briefly review recent work on Siamese networks for object tracking \cite{bertinetto2016fully}. When applying one-shot learning scheme for visual tracking, the object patch in the first frame is used as an exemplar, and the patches in the search regions within the consecutive frames are employed as the candidate instances. The aim is to find the most similar instance from each frame in an embedding space, where the object is represented as a low-dimensional vector. Learning an embedding function with a powerful and discriminative representation is a critical step in this task.

To this end, Bertinetto \textit{et al.} \cite{bertinetto2016fully} apply a deep learning method to learn the embedding function and design a fully convolution Siamese network to reduce computation for real-time speed. This network includes two network branches processing different inputs. One is the exemplar branch used to receive the object bounding box in the first frame. The other is the instances branch applied to process the patches in searching regions of the following frames. These two network branches share the parameters, thus they can be seen as an identical transformation $\phi$ for different inputs. Accordingly, the similar function for an exemplar $z$ and an instance $x$ is defined as $f(z,x) = g(\phi(z),\phi(x))$, where $g$ is a simple similarity metric such as vectorial angle and cross-correlation. In \cite{bertinetto2016fully}, they use the cross-correlation for $g$, and the formulation of function $f$ is transferred as follows:
\begin{equation}
\label{equ_f}
f(z,x) = \phi(z)*\phi(x)+b.
\end{equation}

Then, a logistical loss is applied to the pair-wise loss function, which is formulated as follows:
\begin{equation}
\label{equ_lp}
L_p(y,v) = \sum_{u \in \mathcal{D}}\log(1+e^{(-y[u]\cdot v[u])}).
\end{equation}
where $\mathcal{D}$ is the set of inputs for instances branch, $y[u]\in \{+1,-1\}$ is the ground-truth label of a single exemplar-instance pair $(z,u)$, $v[u]$ is the similarity score of $(z,u)$ i.e. $v[u] = f(z,u)$.

\section{Quadruplet Network}\label{sec:quad}

A quadruplet network (inspired by Siamese networks \cite{bromley1993signature,bertinetto2016fully} and triplet networks \cite{hoffer2015deep}) contains four branches of the same network with shared parameters, while building a direct connection between them through a common loss function. Our proposed quadruplet network structure is shown in Fig. \ref{net}.

\subsection{Network branches}

These four network branches are respectively named as exemplar, instances, positive, and negative branches according to their inputs. We first introduce the input of exemplar branch since other inputs are depended on it. An instance is randomly chosen as an exemplar $z$. Then the instances with the same class are denoted as positive instances and the others are denoted as negative instances. The inputs of instances branches (denoted as a set $\mathcal{D}$) consist of some positive instances and negative instances, which are used for two aspects.

Firstly, these instances are used to learn an function of similarity metric $f(z,x)$ that compares an exemplar $z$ to an instance $x$ and returns a high score with a positive instance and a low score with a negative instance. Secondly, according to these scores, we select the instance with the lowest score from positive instances as the input of positive branch $z_+$ and select the one with the highest score from negative instances as the input of negative branch $z_-$. $z_+$ and $z_-$ have the strongest representation among the inputs of instances branch, since they are on the boundaries of classification in current small sample space (defined on the set $\mathcal{D}$). If we decrease the distance between $z_+$ and $z$, and increase it between $z_-$ and $z$, the corresponding positive boundary will be closer to $z$ and negative one will keep away from $z$. In other words, the margin between boundaries of classification will be increased and it will reduce the error of classification. To achieve this purpose, we design a new loss function for these three branches. More details are shown in the next subsection.

\subsection{Loss function}
The loss function of a quadruplet network consists of two kinds of loss function. The first one is constructed between the outputs of exemplar branch and instances branch, and the second one connects positive, negative and exemplar branches. To give the definition of these two loss functions, we first simplify the notation of branches. In fact, these 4 branches are an identical transformation to different inputs. Thus, we denote them as a transformation function $\phi$. Then the function $f$ is transfered as $f(z,x) = g(\phi(z),\phi(x))$, where $g$ is a simple similarity metric such as vectorial angle and cross correlation. In this paper, we use the cross correlation for $g$ as the same as Siamese network. And the formulation of function $f$ is defined as the same as the function in Eq. (\ref{equ_f}).

Then, we apply a weighted averaged logistical loss to the first pair-wise loss function, which is formulated as follows:
\begin{equation}
\label{equ_l1}
L_1(y,v) = \sum_{u \in \mathcal{D}}w[u]\log(1+e^{(-y[u]\cdot v[u])}).
\end{equation}
where $\mathcal{D}$ is the set of inputs for instances branch, $y[u]\in \{+1,-1\}$ is the ground-truth label of instance $u$, $+1$ and $-1$ represent the labels of positive instance and negative instance, respectively,
$v[u]$ is the similarity score of $(z,u)$ i.e. $v[u] = f(z,u)$, $w[u]$ is the weight for an instance $u$, and $\sum_{u\in \mathcal{D}}w[u]=1, w[u]>0,  u\in \mathcal{D}$. $w[u]$ is a fixed weigh during the general training process, while it may be changed in the pre-computation process.

The pre-computation process means we  first do forward computation once to get the similarity score of each exemplar instance,
which is different from general training method for deep learning.
If the similarity score of a negative instance is larger than the minimal score of positive instances, it will violate the assumption of similarity, i.e. the score of positive instance should be larger than negative instance. Thus we strengthen the punishment for this situation by increasing the corresponding weights.
For any case that violates the assumption, we enlarge the weight parameter $w[u]$ by a factor of 2, i.e. $w_{new}[u] = 2 w_{original}[u]$.
Then, these adapted weights are applied for forward computation and backward computation to finish a training process.

Inspired by the triplet network \cite{hoffer2015deep}, the second loss function is constructed by a mean square error on the soft-max results of similarity scores between different branches. Comparing soft-max results to $(0,1)$ vector, we can get the triplet loss function:
\begin{equation}
\label{eq:l2}
L_2(s_+,s_-) = \|(s_+-1,s_-)\|^2_2,
\end{equation}
where
\begin{equation}
\label{eq:s_p}
s_+ = \frac{e^{f(z,z_+)}}{e^{f(z,z_+)}+e^{f(z,z_-)}}
\end{equation}
and
\begin{equation}
\label{eq:s_n}
s_- = \frac{e^{f(z,z_-)}}{e^{f(z,z_+)}+e^{f(z,z_-)}}.
\end{equation}

\begin{figure*}
  \centering
   \includegraphics[width = .95 \textwidth]{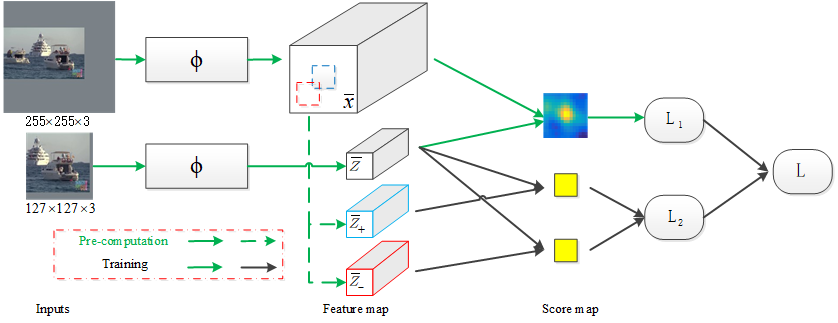}
  \caption{The framework of a single training iteration during the tracking process, which consists of a precomputing phase and a training phase. The green arrows represent the precomputation, which aims to select the powerful positive and negative feature maps ($\bar{z}_+$ and $\bar{z}_-$) from feature map of instances $\bar{x}$. They produce the adapted weights for pairs loss $L_1$. The solid arrows indicate the training phase include the forward and backward computations. The definitions of $\phi,L_1,L_2,L$ are the same as Fig. \ref{net}.
  {Note that the $L_1$ loss is derived from exemplar and instances, while $L_2$ is from exemplar, positive instance and negative instance. $L$ loss is the weighted combination of these two losses.}
}
  \label{train_track}
\end{figure*}

The final loss function is the weighted sum of these two loss functions:
\begin{equation}
L = \frac{1}{\sum_{i=1}^2\bar{w}[i]}\sum_{i=1}^2 \bar{w}[i]L_i,
\end{equation}
where $\bar{w}[1],\bar{w}[2]>0$ are the weights to balance the loss $L_1$ and $L_2$. They are not prior constants but learned during training.

The parameters of branch network $\theta$ and the weights of loss $\bar{w}$ are obtained by applying Stochastic Gradient Descent (SGD) to the problem:
\begin{equation}
\arg \min_{\theta,\bar{w}}\quad \mathbb{E}\quad L(z,z_+,z_-,\mathcal{D},y;\theta,\bar{w}).
\label{eq_L}
\end{equation}

For our neural networks, we propose three new layers using $L_1$, $L_2$ and $L$ losses. The backward computation of $L_1$ loss is similar to the logistic loss since the weights are fixed during backward computing. We only need to add the weights into the derivatives of logistical loss. The other two are not similar to the common loss. Thus, we give the gradients of these two losses for backward computation. 
$L_2$ loss layer can be decomposed as a square error layer (Eq. \ref{eq:l2}) and a soft-max layer (Eq. \ref{eq:s_p}, \ref{eq:s_n}).
We get the derivatives of the square error as:
\begin{equation}
\frac{\partial L_2}{\partial s_+} = 2(s_+-1)=-2s_-, \qquad  \frac{\partial L_2}{\partial s_-} = 2s_-.
\end{equation}

To avoid too big gradient in soft-max layer, we reformulate Eq. \ref{eq:s_p} as follows:
\begin{equation}
s_+  = \frac{e^{f_+}/e^m}{(e^{f_+}+e^{f_-})/e^m} = \frac{e^{f_+-m}}{e^{f_+-m}+e^{f_- - m}}.
\end{equation}

Similarly, the reformulation of Eq. \ref{eq:s_n} is
\begin{equation}
s_-  = \frac{e^{f_--m}}{e^{f_+-m}+e^{f_- - m}}.
\end{equation}
where $f_+$ and $f_-$ are similarity scores $f(z,z_+)$ and $f(z,z_-)$, $m$ is the maximal between $f_+$ and $f_-$ i.e. $m = max(f_+,f_-)$.
{In fact, the maximal $m$ is used to avoid numeric overflow since the output of exponential function is very easy to overflow.}

Then we can get Jacobian matrix $\mathbf{J}$:
\begin{equation}
\mathbf{J} =
\left[\begin{matrix}
\frac{\partial s_+}{\partial f_+} & \frac{\partial s_+}{\partial f_-}\\
\frac{\partial s_-}{\partial f_+} & \frac{\partial s_-}{\partial f_-}
\end{matrix}\right]
=
\left[\begin{matrix}
s_+(1-s_+) & -s_+s_-\\
-s_+s_- & s_-(1-s_-)
\end{matrix}\right]
\end{equation}

According to the chain rule, we can get the partial derivatives as follows:
\begin{equation}
\left[\begin{matrix}
\frac{\partial L_2}{\partial f_+} \\
\frac{\partial L_2}{\partial f_-}
\end{matrix}\right]
= \mathbf{J}^T
\left[\begin{matrix}
\frac{\partial L_2}{\partial s_+} \\
\frac{\partial L_2}{\partial s_-}
\end{matrix}\right]
=
\left[\begin{matrix}
-4 s_+ s_-^2 \\
4 s_+ s_-^2
\end{matrix}\right]
\end{equation}

The loss layer $L$ contains two inputs ($L_1$ and $L_2$) and two parameters ($\bar{w}[1]$ and $\bar{w}[2]$),
and its partial derivatives are formulated as:
\begin{equation}
\frac{\partial L}{\partial L_j} = w[j],
\end{equation}
and
\begin{equation}
\begin{aligned}
\frac{\partial L}{\partial \bar{w}[j]} & = \frac{\partial L}{\partial \bar{w}[j]}  ( \frac{1}{\bar{w}_s} ) (\sum_{i=1}^2 \bar{w}[i]L_i) +(\frac{1}{\bar{w}_s} ) \frac{\partial L}{\partial \bar{w}[j]} (\sum_{i=1}^2 \bar{w}[i]L_i)  \\
									   & =	-\frac{1}{\bar{w}_s^2}\sum_{i=1}^2 \bar{w}[i]L_i +\frac{1}{\bar{w}_s}L_j\\
	& = \frac{1}{\bar{w}_s^2}(\bar{w}_s L_j-\sum_{i=1}^2  \bar{w}[i]L_i),\\
\end{aligned}
\end{equation}
where $j=1,2$ and $\bar{w}_s = \sum_{i=1}^2  \bar{w}[i]$.
In practice, we set a small threshold $T$, such as 0.01, for $\bar{w}[i]$ to insure $\bar{w}[i]>0$ by imposing $\bar{w}[i] = \max(T,\bar{w}[i])$.
{As shown in Fig. \ref{train_track}, we directly use the feature maps $\bar{Z}_+$ and $\bar{Z}_-$ obtained from pre-computation as the inputs of positive and negative braches. Thus, their partial derivatives are not applied to update the CNN network $\phi$. Only the partial derivatives of exemplar feature $\bar{Z}$ are used for training during backward computation to reduce time cost.}

\begin{figure*}
  \centering
   \includegraphics[width = .88 \textwidth]{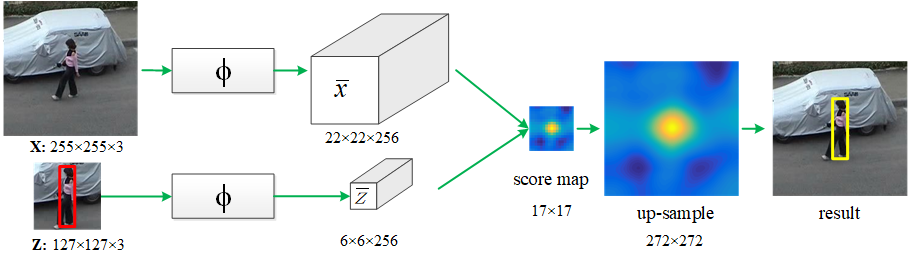}
  \caption{The framework of our online tracking method. Above, $z$ and $x$ represent the input exemplar image and the search image, and $\bar{z}$ and $\bar{x}$ are the corresponding embedding features. The score map is calculated by the similarity between the sub-windows of $\bar{x}$ and $\bar{z}$. After upsampling the score map, we find the location of the object in the search image according to the location of maximum score in the upsampled map.}
  \label{test_track}
\end{figure*}

\subsection{Framework of our tracker}
We apply the proposed Quadruplet network on an one-shot learning problem in visual object tracking. Given an object of interest on the first video frame, the task of single object tracking is to find the object on the following frames. In the opinion of one-shot learning, a sub-window enclosing the object is seen as an exemplar and the sub-windows of each frame are viewed as a candidate of instances. The goal is to find the most similar one with exemplar in each frame. Our tracking method consists of off-line training and online testing stages using different network structures.

In the off-line training stage, we use quadruplet network to achieve powerful embedding representation. The architecture of shared convolutional network is the same as SiamFc-3s \cite{bertinetto2016fully}, which is a variation of the network of Krizhevsky \textit{et al.} \cite{krizhevsky2012imagenet}.

As shown in Fig. \ref{train_track}, we simplify the quadruplet network by selecting powerful feature patches at the last convolutional layer of instances branch instead of positive and negative branch. Pairs of an exemplar image and a larger search image are applied to inputs of exemplar and instances branch, where each sub-window of the same size with the exemplar image is an candidate instance. The scores $v$ in equation (\ref{equ_l1}) will become a score map as shown in Fig. \ref{train_track}. Its label map is designed according to the location i.e. we set positive label $+1$ to points in the central area within radius $R$ ($R=2$ in our experiments), which is denoted as positive area $\mathcal{D}_p$, and ones in other area (negative area $\mathcal{D}_n$) are set as $-1$. Before each training iteration, we first precompute the score map, and respectively select the instance in the center and the one with the highest score in negative area as the powerful positive and negative instance. Their last convolutional features are set as the inputs of triplet loss. Otherwise, the precomputed score map is also used to construct the weighted pair loss. The initial weights are defined as balance weights to balance the number of positive and negative instances. The formulation is defined as:
\begin{equation}
  w[u]=\begin{cases}
    \frac{1}{2|\mathcal{D}_p|}, & u\in  \mathcal{D}_p\\
    \frac{1}{2|\mathcal{D}_n|},  & u\in \mathcal{D}_n\\
  \end{cases}
\end{equation}

If a negative score is more than the minimal of positive scores, we increase the weight as $w[u] = 2w[u]$ and then normalize all the weights as $w[u]=w[u]/\sum w[u]$. After precomputation, we do forward and backward computation to finish a training iteration.

\begin{figure*}
  \centering
	\includegraphics[width = .96\textwidth]{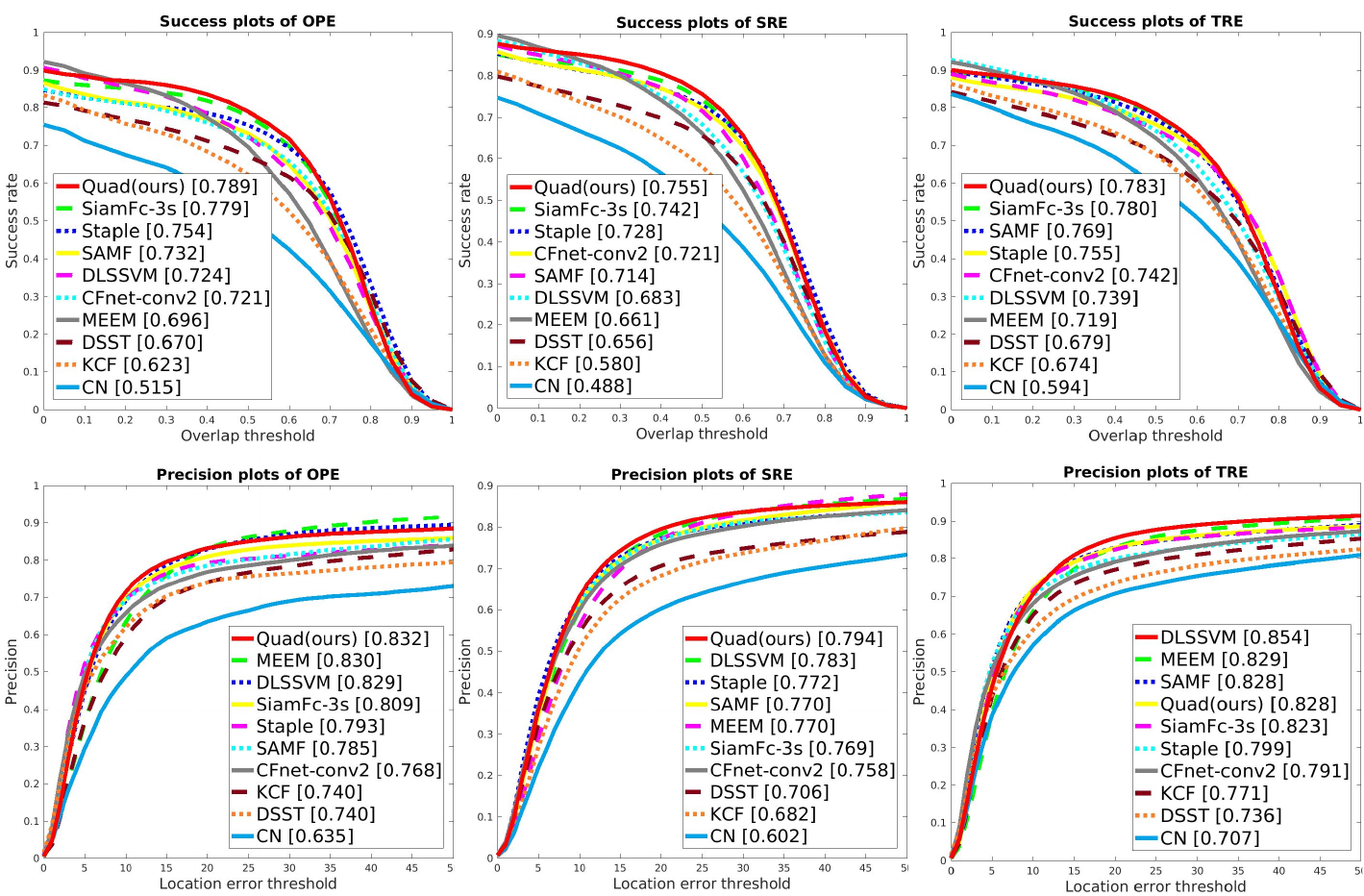}
  \caption{The comparison results of OTB-2013 \cite{wu2013online} benchmark. Success and precision plots for OPE, SRE and TRE which are one pass evaluation, spatial robustness evaluation, and temporal robustness, respectively.
  }
  \label{res_otb}
\end{figure*}

In the online testing (tracking) phase, only the exemplar branch and the instances branch are used. We crop an exemplar image in the first frame and also enlarge the searching images (searching region) in the consecutive frames. The search images center on the location of the object in the previous frames. The exemplar and the search images are resized to $127\times127$ and $255\times255$, respectively. Using these inputs, our tracking network calculates a score map as shown in Fig. \ref{test_track}. Then, the score map is upsampled by bicubic interpolation from $17\times17$ to $272\times272$ to achieve a higher location accuracy. The location of the object is determined by the maximum score in the upsampled map.

\section{Experimental results}\label{sec:exp}
\subsection{Implementation details}
\textbf{Training.} We use MatConvNet \cite{vedaldi2015matconvnet} to train the parameters of the shared network $\theta$ and the weights of loss $\bar{w}$ by minimizing Eq. (\ref{eq_L}) with SGD. The initial values of the shared network are copied from the trained model in SiamFc-3s \cite{bertinetto2016fully} and the weights are set as $(0.9,0.1)$. We use the same training and validation sets with \cite{bertinetto2016fully}. They are randomly sampled from the `new object detection from video challenge' in the 2015 edition of the ImageNet Large Scale Visual Recognition Challenge [10] (ILSVRC). The dataset contains almost 4500 videos with 30 different classes of animals and vehicles. Training is performed over 10 epochs, each consisting of 53,200 sampled pairs. We randomly select $10\%$ pairs as the validation set at each epoch, and the final network used for testing is determined by the minimal mean error of distance (presented in \cite{bertinetto2016fully}) on the validation set. The gradients for each iteration are estimated using mini-batches of size 8, and the learning rate is decayed geometrically after epoch from $10^{-2}$ to $10^{-5}$. To handle the gray videos in benchmarks, $25\%$ of the pairs are converted to grayscale during training.

\textbf{Tracking.} As mentioned before, only the initial object is selected as the exemplar image. Thus, we compute the embedding feature $\phi(z)$ once and compare it with the searching images of the subsequent frames. To handle scale variations, three scales $1.0375^{\{-1,0,1\}}$ are searched for the object, and the scale is updated by linear interpolation with a factor of 0.59 to avoid huge variation of scale. Our Intel Core i7-6700 at 3.4 GHz machine is equipped with a single NVIDIA GeForce 1080, and our online tracking method runs at 78 frames-per-second.

\begin{figure*}
  \centering
    \includegraphics[width = .96\textwidth]{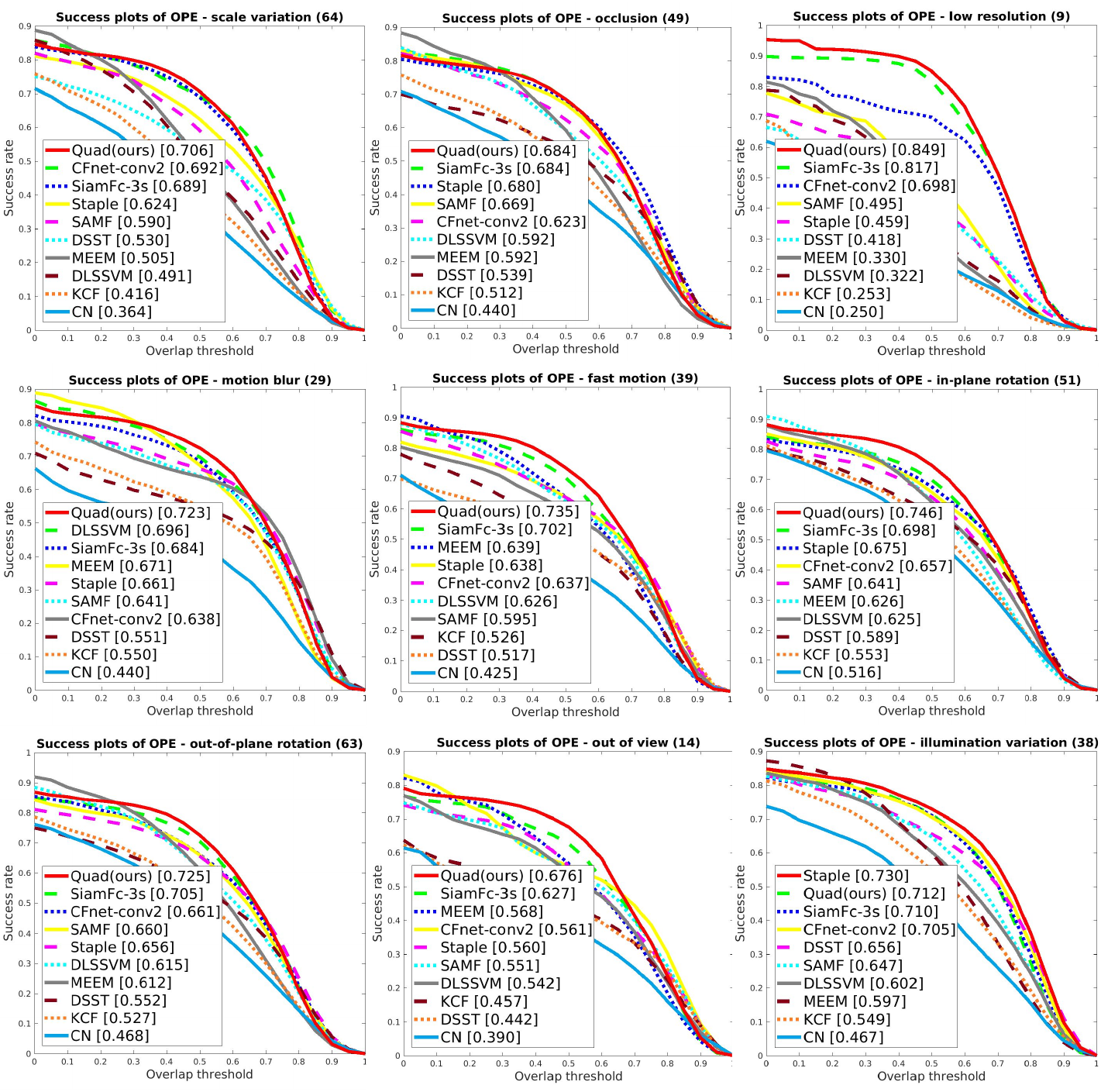}    
  \caption{Overlap success plots of OPE with AUC for 9 tracking challenges on OTB-100.}
  \label{res_att}
\end{figure*}

\subsection{Benchmarks and evaluation metric}
We evaluate our tracking method, which we call as `Quad' with the recent state-of-the-art trackers in popular benchmarks including OTB-2013 \cite{wu2013online}, OTB-50, OTB-100 \cite{wuobject2015}, and VOT-2015 \cite{Kristan2015a}.

The OTB-2013 benchmark contains 50 challenging sequences and uses different metrics
to evaluate tracking methods. The authors expand the OTB-2013 to OTB-100 including 100 sequences and select 50 more challenging sequences denoted OTB-50 as a small benchmark. In this paper, we use the overlap success rate and distance precision metrics \cite{wu2013online} to evaluate trackers on OTB-2013, OTB-50, and OTB-100. Overlap success rate measures the intersection-over-union (IoU) of ground-truth and predicted bounding boxes. The success plot shows the rate of bounding boxes whose IoU score is larger than a given threshold. We apply the overlap success rate in terms of threshold 0.5 to rank the trackers. The precision metric means the percentage of frame locations within a certain threshold distance from those of the ground truth. The threshold distance is set as 20 for all trackers. The VOT-2015 benchmark is a challenging dataset to evaluate the short-term tracking performance since in VOT-2015, a tracker is restarted in the case of a failure, where there is no overlap between the predicted bounding box and ground truth. It contains 60 sequences collected from some previous tracking benchmarks (356 sequences in total). For this dataset, we evaluated tracking performance in terms of accuracy (overlap with the ground-truth), robustness (failure rate), and Expected Average Overlap (EAO which is principled combination of accuracy and robustness) \cite{Kristan2015a}.

\subsection{The OTB-2013 benchmark}
On OTB-2013 benchmark, we compare our Quad tracker against several state-of-the-art trackers that can operate in real-time: CFnet-conv2 \cite{valmadre2017end}, SiamFc-3s \cite{bertinetto2016fully}, Staple \cite{bertinetto2016staple}, CN \cite{danelljan2014accurate}, and KCF \cite{Henriques2015High}. For reference, we also compare with recent trackers: DSST \cite{danelljan2014accurate}, MEEM \cite{Zhang2014MEEM}, SAMF \cite{li2014scale}, DLSSVM \cite{ning2016object}.

\textbf{Overall comparison.} A conventional way to evaluate trackers is one-pass evaluation (OPE), however some trackers may be sensitive to the initialization. To measure the performance with different initialization,
we use spatial robustness evaluation (SRE) and temporal robustness evaluation (TRE). SRE uses different bounding boxes in the first frame and TRE starts at different frames for initialization.
In Fig. \ref{res_otb}, our method outperforms over recent state-of-the-art real time trackers in terms of overlap success rate and precision for OPE, SRE and TRE, respectively. Compared with our baseline SiamFc-3s, the results show that our training method is able to achieve more robust and powerful feature for tracking. Among referenced trackers, our tracker also performs better in five evaluation metric except the precision for TRE. In this metric, our Quad method achieve the 4th rank (0.828), which is very close to the 3rd SANF (0.828) and the 2nd MEEM (0.829). SAMF ranks 1st in precision while ranks 6th in success for TRE.

\subsection{Results on OTB-50 and OTB-100}
On OTB-50 and OTB-100 benchmarks, we also compare the recent trackers mentioned on OTB-2013 comparison.

\begin{figure*}
  \centering
	\includegraphics[width = .96\textwidth]{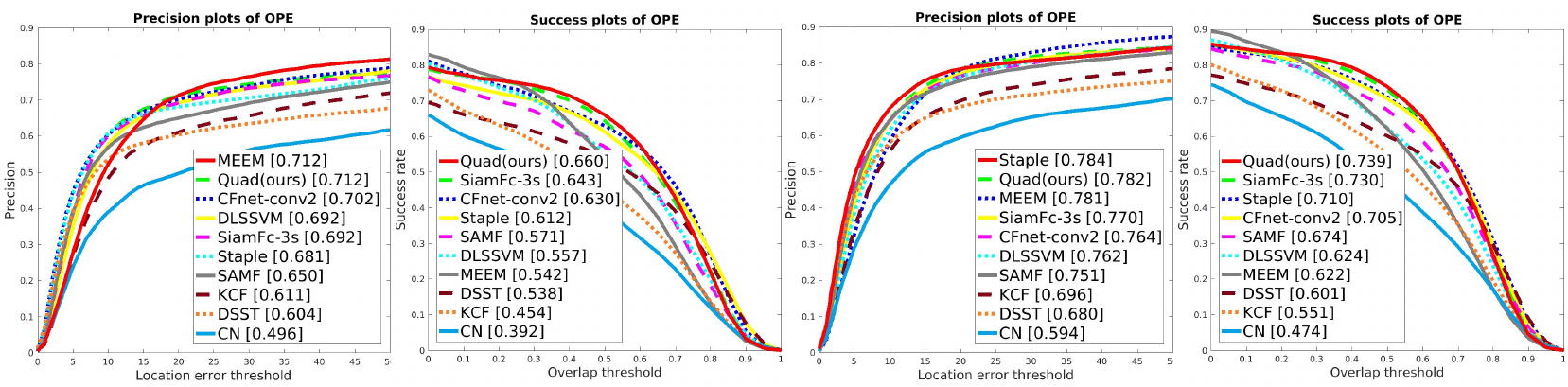}
	
        (a) OTB-50 ~~~~~~~~~~~~~~~~~~~~~~~~~~~~~~~~~~~~~~~~~~~~~~~~~~~~(b) OTB-100

  \caption{Precision and success plots with AUC for OPE on OTB-50 and OTB-100 \cite{wuobject2015} benchmark.}
  \label{res_otb50_100}
\end{figure*}

\begin{figure*}
  \centering
    \includegraphics[width = .98 \textwidth]{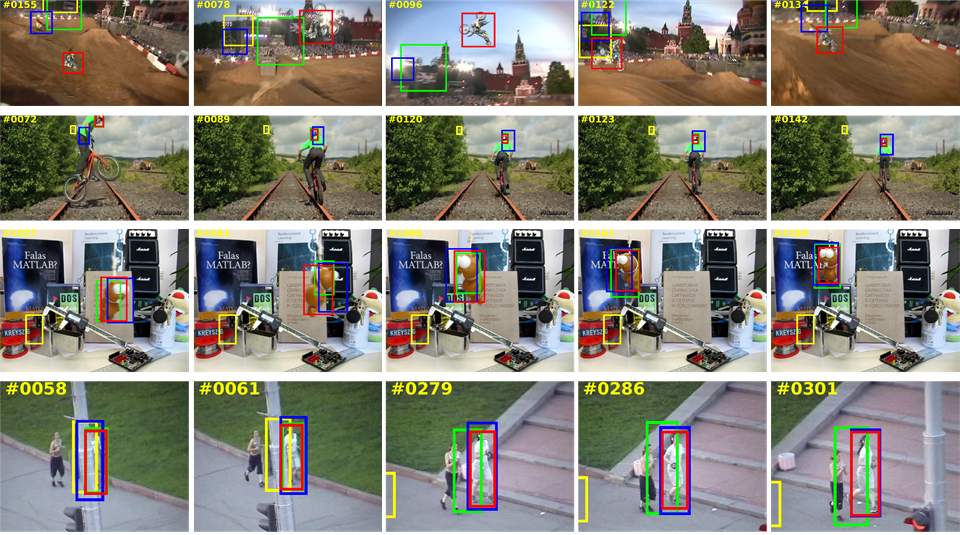}
    Staple \textcolor{yellow}{ \rule[0.3ex]{10mm}{1mm}}  CFnet-conv2 \textcolor{blue}{ \rule[0.3ex]{10mm}{1mm}} SiamFc-3s \textcolor{green}{ \rule[0.3ex]{10mm}{1mm}}    Quad (Ours) \textcolor{red}{ \rule[0.3ex]{10mm}{1mm}}
  \caption{Qualitative evaluation of recent real-time trackers
Staple \cite{bertinetto2016staple}, CFnet-conv2 \cite{valmadre2017end}, SiamFc-3s \cite{bertinetto2016fully} and our algorithm Quad. Nine challenging sequences (from top to down) are MotorRolling, DragonBaby, Skiing, KiteSurf, Biker, Lemming, Coke, Jogging2, Jumping, respectively) are compared among the state-of-the-art trackers.}
  \label{res_quality}
\end{figure*}

\textbf{Attribute-based Performance Analysis.} In OTB-100 benchmark, the sequences are annotated with 11 attributes for different challenging  factors including Illumination Variation (IV), Scale Variation (SV),  Occlusion (OCC), Deformation (DEF), Motion Blur (MB), Fast Motion (FM), In-Plane Rotation (IPR), Out-of-Plane Rotation (OPR), Out-of-View (OV), Background Clutters (BC), and Low Resolution (LR).
To evaluate the proposed method in terms of each challenging factors, we compare our method with other trackers with different dominate attributes. Fig. \ref{res_att} shows the results of 9 main challenging attributes evaluated by the overlap success plots of OPE. 
Our approach outperforms other trackers in 8 subsets: SV, OCC, LR, MB, FM, IPR, OPR, and OV, especially in LR and IPR.
In other subset, Staple performs the best while our approach gets the second best performance. 
This tracker is based on correlation filter and apply Fourier transform. Maybe adding the Fourier transform into our method will further improve the performance. Compared with baseline SiamFc-3s, our tracker outperforms it in all subsets.

\begin{table*}
\caption{ Evaluation on VOT2015 by EAO, the weighted means of accuracy, robustness and speed. We run our method and SiamFc-3s \cite{bertinetto2016fully}, and report their speed (FPS). Otherwise (*) we report the values from the VOT2015 results \cite{Kristan2015a} in EFO units, which roughly correspond to FPS (e.g. the speed of the NCC tracker is 140 FPS with 160 EFO). The \textcolor{red}{first} and  \textcolor{green}{second} best scores are highlighted in color.}
\centering
\begin{tabular}{cccccccccccc}
\hline
     & Quad(ours) & SiamFc-3s & BDF   & NCC    & FOT   & ASMS  & FCT   & matFlow & SKCF  & PKLTF & sumShift \\
     \hline
EAO  & \textcolor{red}{0.261}      & \textcolor{green}{0.248}     & 0.153 & 0.080  & 0.139 & 0.212 & 0.151 & 0.150   & 0.162 & 0.152 & 0.234    \\
Acc. & \textcolor{red}{0.553}      & \textcolor{green}{0.550}     & 0.401 & 0.500  & 0.432 & 0.507 & 0.431 & 0.420   & 0.485 & 0.453 & 0.517    \\
Rob. & \textcolor{green}{1.791}      & 1.818     & 3.106 & 11.345 & 4.360 & 1.846 & 3.338 & 3.121   & 2.681 & 2.721 & \textcolor{red}{1.682}    \\
FPS  & 78         & 78        & \textcolor{red}{175*}  & \textcolor{green}{135*}   & 126*  & 101*  & 73*   & 71*     & 58*   & 26*   & 15*    \\
\hline
\end{tabular}
\label{table:vot2015}
\end{table*}

\begin{figure*}
  \centering
    \includegraphics[width = .78 \textwidth]{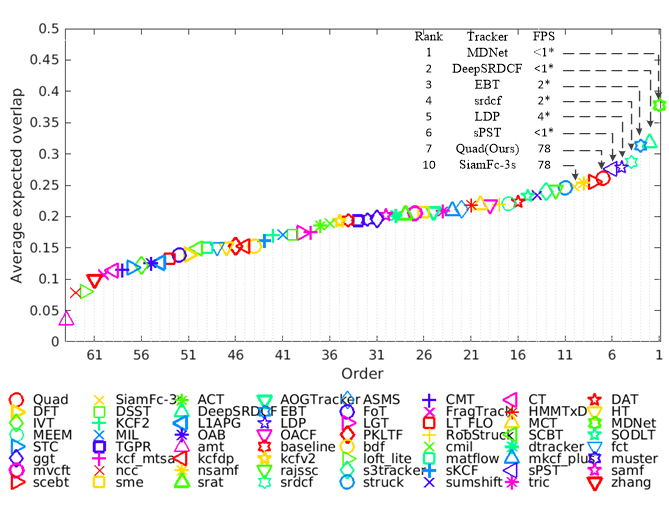}
  \caption{Expected Average Overlap plot for our tracker, SiamFc-3s \cite{bertinetto2016fully}, and the total 62 trackers in VOT-2015 \cite{Kristan2015a}. We show
  the speeds of our tracker, SiamFc-3s \cite{bertinetto2016fully}, and the top 6 trackers from VOT-2015: MDNet \cite{nam2015learning}, DeepSRDCF \cite{danelljan2015learning}, EBT \cite{zhu2015tracking}, srdcf \cite{danelljan2015learning}, LDP \cite{Kristan2015a}, and sPST \cite{hua2015online}.
  Here the symbol (*) represents the approximate fps of a tracker according to its EFO units in VOT-2015 results.}
  \label{res_vot2015}
\end{figure*}

\textbf{Overall comparison.} Both precision and success metrics are reported for OPE. Fig. \ref{res_otb50_100} shows that our tracker also achieves improvement compared with our baseline SiamFc-3s in these two benchmarks in terms of precision and success metrics. In success metric, our method performs better than all trackers on these two benchmarks. In precision metric, our tracker achieves the second best performance on OTB-50 following the first one MEEM with slightly reducing from 0.7122 to 0.7117, while we get significant improvement from 0.542 to 0.660. Similarly, on OTB-100, our tracker ranks second (0.782) slightly lower than the first one Staple (0.784) in precision while increases the success rate from 0.710 to 0.739.

\subsection{Qualitative Evaluation}
We compare our Quad method with other three state-of-the-art real-time trackers (Staple \cite{bertinetto2016staple}, CFnet-conv2 \cite{valmadre2017end}, SiamFc-3s \cite{bertinetto2016fully}) on four challenging sequences from OTB-100 \cite{wuobject2015} in Fig. \ref{res_quality}.
In the most challenging sequences such as MotorRolling (first row), most methods fail to track objects well,
whereas our Quad algorithm performs accurately in terms of either precision or overlap.
The Staple, CFnet-conv2, SiamFc-3s methods lose track of the target gradually due to significant deformation and fast motion in Biker sequences.
Lemming sequences demonstrate the performance of these trackers in handling scenarios with rotation and occlusion, only our tracker succeeds throughout this whole sequence. The Staple, SiamFc-3s trackers are not able to keep tracking the target after occlusion and deformation in Jogging2 sequence.
Overall, the proposed Quad tracker is able to get powerful and robust features to handle these challenges well to alleviate tracker drift.

\subsection{Evaluation on VOT-2015}
In our experiments, we use the Visual Object Tracking 2015 (VOT-2015) toolkit, which contains the evaluation in short-term visual object tacking tasks.

\textbf{Fast speed}: We compare our tracker with SiamFc-3s \cite{bertinetto2016fully} and 9 top participants in the VOT-2015 in terms of speed, including BDF \cite{maresca2014clustering}, FOT \cite{vojivr2014enhanced}, ASMS \cite{vojir2014robust}, NCC, FCT, matFlow, SKCF, PKLTF \cite{Kristan2015a}, and sumShift \cite{lee2011visual}.
Table \ref{table:vot2015} shows that our tracker achieves the best Expected Average Overlap (EAO)
and the highest accuracy among the most accurate trackers with speed more than 15 fps.
Among the fast trackers, the highest robustness (1.682) belongs to sumShift followed by ours (Quad) (1.791). Our tracker significantly improves the accuracy and robustness of most participants with top speed in VOT-2015 (BDF, FOT, ASMS, NCC, FCT, matFlow, SKCF, PKLTF) and SiamFc-3s.

\textbf{Overall}: For reference, we also compare our approach with SiamFc-3s and total 62 trackers in VOT-2015. In Fig. \ref{res_vot2015}, our tracker ranks at 7th/64 in terms of EAO, following MDNet \cite{nam2015learning}, DeepSRDCF \cite{danelljan2015learning}, EBT \cite{zhu2015tracking}, srdcf \cite{danelljan2015learning}, LDP \cite{Kristan2015a}, and sPST \cite{hua2015online}. These 6 trackers  take a lot of time to process each frame to achieve better performance. Thus, they operate on low frame-rate (less than 5 fps). In contrast, our tracker achieves high speed (78 fps) beyond real-time further faster than them. Finally, it is worth to mention that the baseline SiamFc-3s ranks 10th while our approach also performs better than it.

\begin{figure}
	\centering
	\includegraphics[width = .24\textwidth]{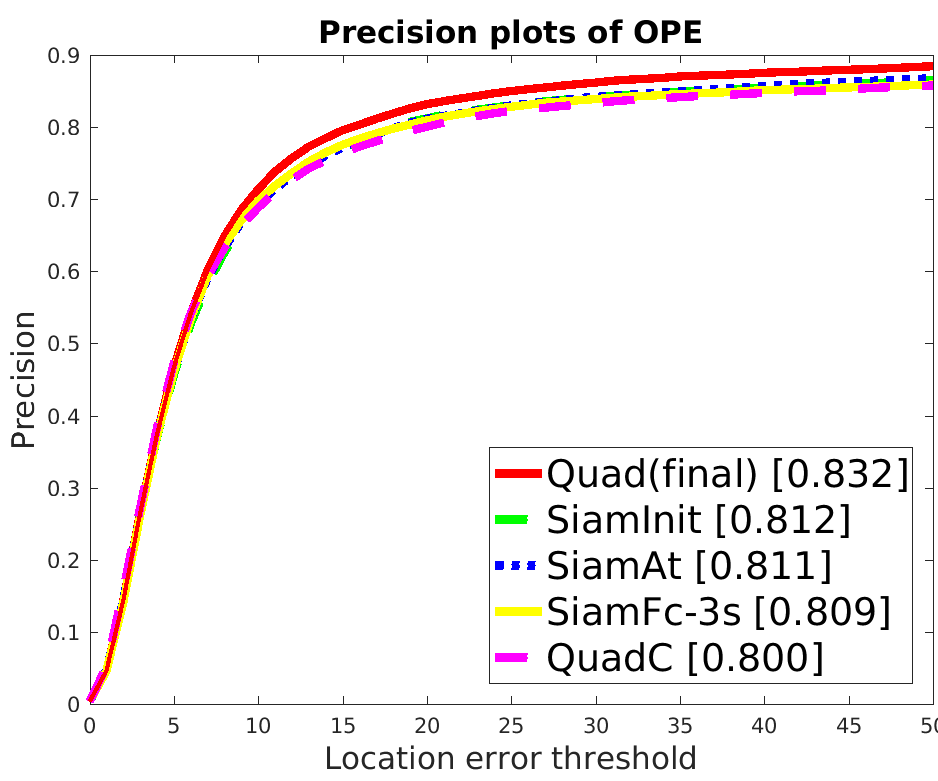}
	\includegraphics[width = .24\textwidth]{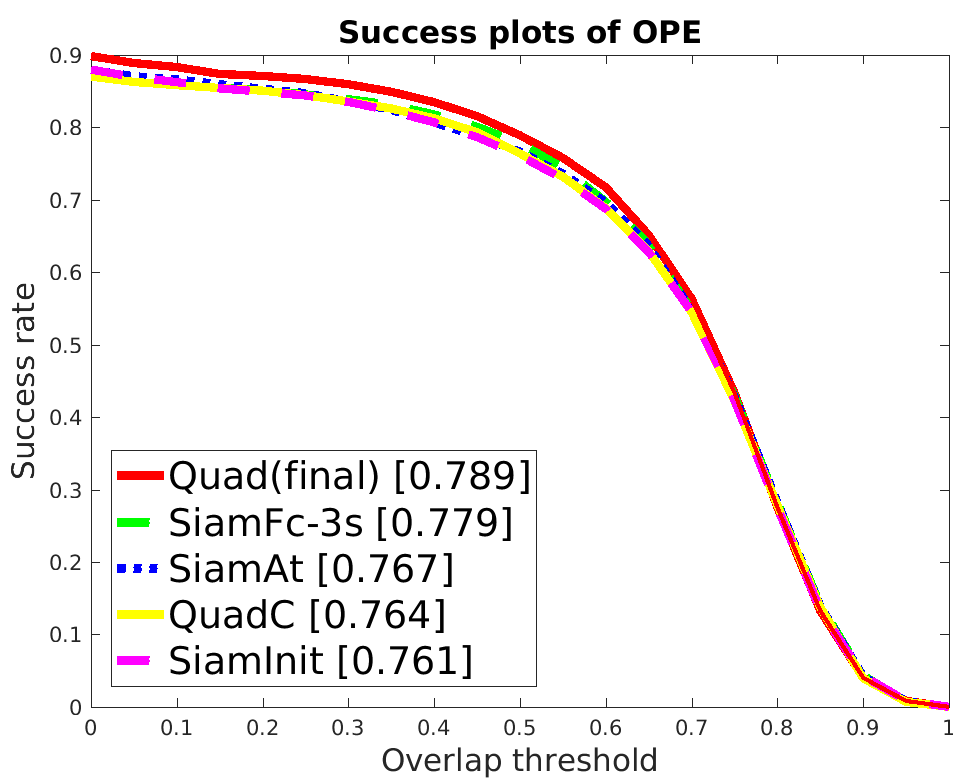}
	\caption{Comparison results of self-comparison with different variants of our tracker. The plots show the precision and overlap success rate on OTB-2013 \cite{wu2013online} in terms of OPE.}
	\label{res_self}
\end{figure}
\subsection{Ablation Study}
To evaluate the effect of different components in our method, we compare the tracker SiamFc-3s \cite{bertinetto2016fully} against different variants of our tracker: SiamInit, SiamAt, QuadC, and Quad. SiamFc-3s can be seen as the baseline of our method. Since our online tracking method is the same with it. The difference is the training model. SiamInit is the variant of SiamFc-3s, which is initialized with the final network parameters in SiamFc-3s and trained again over 10 epochs. SiamAt is also a Siamese network but trained with the proposed adaptive weighted pairs loss. QuadC is the version that combines the weighted pairs loss and the triplet loss with constant weights $\bar{w}=[0.9,0.1]$, and Quad is the final version with the learned weights.

We evaluate these trackers with one-pass evaluation (OPE) while running them throughout a test sequence with an initialization from the ground-truth position in the first frame. In Fig. \ref{res_self}, directly training more epochs (SiamInit) will improve precision but reduce overlap success rate compared with the baseline SiamFc-3s. This indicates that representation power of the original network was close to its limit. More training may not improve more performance. Thus, we seek for an alternative solution for improvement. Firstly, we try an adaptive weighted pairs loss and achieve slightly overall improvement with increasing overlap, while slightly reducing precision compared with SiamInit. Secondly, we want to add triplet loss to mine the potential relationships among samples further. However, directly adding triplet loss into a Siamese net with prior weights may reduce the performance (see QuadC in Fig. \ref{res_self}) since the prior weights may not be optimal for the combination of triplet loss and pair loss. To solve this problem, we design a weight layer to choose suitable combination weights during training and achieve better precision and overlap success rate as shown with Quad in Fig. \ref{res_self}.

\section{Conclusion}\label{sec:conclusion}
We demonstrated that the proposed quadruplet network using multi-tuples for training allows accurate mining of the potential connections among instances and achieves more robust representations for one-shot learning. We have shown that by automatically adjusting the combinational weights between triplet and pair loss, we improve the training performance. We analyzed the feasibility of our quadruplet network in visual object tracking. Our results indicate that our tracking method outperforms others while maintaining beyond real-time speed. As future work, we employ different loss functions and extend the quadruplet networks to other computer vision tasks.



\begin{IEEEbiographynophoto}
{Xingping Dong} is currently working toward the Ph.D. degree in the School of Computer Science, Beijing Institute of Technology, Beijing, China.
His current research interests include video tracking and deep learning.
\end{IEEEbiographynophoto}
\vspace{-6mm}
\begin{IEEEbiographynophoto}
{Jianbing Shen} (M'11-SM'12) is a Professor with the School of
Computer Science, Beijing Institute of Technology, Beijing, China.
He is also acting as the Lead Scientist at the Inception Institute of Artificial Intelligence, Abu Dhabi, United Arab Emirates.
He received his Ph.D. from the Department of Computer Science, Zhejiang University in 2007.
He has published more than 100 top journal and conference papers, six papers are selected as the ESI Hightly Cited or ESI Hot Papers.
His current research interests are in the areas of deep learning for video analysis, computer vision for autonomous driving,
deep reinforcement learning, and machine learning for intelligent systems.

Dr. Shen is a Senior Member of \textit{IEEE}. He has also obtained many flagship honors including
the Fok Ying Tung Education Foundation from Ministry of Education, the Program for
Beijing Excellent Youth Talents from Beijing Municipal Education Commission, and the
Program for New Century Excellent Talents from Ministry of Education.
He serves as an Associate Editor for \textit{IEEE Trans. on Neural Networks and Learning Systems},
\textit{Neurocomputing}, \textit{the visual computer} and other journals.
\end{IEEEbiographynophoto}
\vspace{-6mm}
\begin{IEEEbiographynophoto}
{Dongming Wu} is currently working toward the Master degree in the School of Computer Science,
Beijing Institute of Technology, Beijing, China.
His current research interests include reinforcement learning and object tracking.
\end{IEEEbiographynophoto}
\vspace{-6mm}
\begin{IEEEbiographynophoto}
{Kan Guo} received the Ph.D. degree from Beihang University in 2018.
He is currently a senior algorithm engineer in Taobao Technology Department of Alibaba Group, Hangzhou, China.
His research interests include computer vision, machine learning, and computer graphics.
\end{IEEEbiographynophoto}
\vspace{-6mm}
\begin{IEEEbiographynophoto}
{Xiaogang Jin (M'04)} is a Professor in the State Key Laboratory of CAD$\&$CG, Zhejiang University.
He received the B.Sc. degree in computer science and the M.Sc. and Ph.D degrees
in applied mathematics from Zhejiang University, P. R. China, in 1989, 1992, and 1995, respectively.
His current research interests include image processing, traffic simulation, collective behavior simulation,
cloth animation, virtual try-on, digital face, implicit surface modeling and applications, creative modeling,
computer-generated marbling, sketch-based modeling, and virtual reality.
He received an ACM Recognition of Service Award in 2015 and two Best Paper Awards from CASA 2017 and CASA 2018.
He is a member of the IEEE and ACM.
\end{IEEEbiographynophoto}
\vspace{-6mm}
\begin{IEEEbiographynophoto}
{Fatih Porikli} (M'96, SM'04, F'14) is an IEEE Fellow and a Professor in the Research School of Engineering, Australian National University (ANU).
He is also acting as the Chief Scientist at Huawei, Santa Clara. He has received his Ph.D. from New York University in 2002.
His research interests include computer vision, pattern recognition, manifold learning, image enhancement, robust and sparse optimization and online learning with commercial applications in video surveillance, car navigation, intelligent transportation, satellite, and medical systems.
Prof. Porikli is the recipient of the R\&D 100 Scientist of the Year Award in 2006.
He is serving as the Associate Editor of 5 journals including \textit{IEEE Signal Processing Magazine},
\textit{SIAM Imaging Sciences}, \textit{Machine Vision Applications}, \textit{IEEE Trans. on Multimedia}.
\end{IEEEbiographynophoto}

\vfill
%
\end{document}